\documentclass{article}
\usepackage{spconf,amsmath,graphicx,amssymb, float}
\usepackage{xcolor}
\usepackage[export]{adjustbox}
\usepackage{array,multirow,algorithm,algpseudocode}
\usepackage{siunitx}
\usepackage{hyperref}

\title{Learning with Less data via weakly labeled patch classification in Digital Pathology}
\name{Eu Wern Teh and Graham W.~Taylor}
\address{School of Engineering, University of Guelph, ON, Canada\\
    Vector Institute, ON, Canada}
\begin{document}
%
\maketitle
\begin{abstract}
In Digital Pathology (DP), labeled data is generally very scarce due to the requirement that medical experts provide annotations. We address this issue by learning transferable features from weakly labeled data, which are collected from various parts of the body and are organized by non-medical experts. In this paper, we show that features learned from such weakly labeled datasets are indeed transferable and allow us to achieve highly competitive patch classification results on the colorectal cancer (CRC) dataset~\cite{kather2016multi} and the PatchCamelyon (PCam) dataset~\cite{veeling2018rotation} while using an order of magnitude less labeled data.

\end{abstract}
\section{Introduction}\label{sec:intro}

In recent years, we have witnessed significant success in visual recognition for Digital Pathology~\cite{bejnordi2017diagnostic, aresta2019bach}. The main driving force of this success is due to the availability of labeled data by medical experts. However, creating such datasets is costly in terms of time and money.

\begin{figure}[H]
\centering
{\setlength{\tabcolsep}{2pt}
\scalebox{0.75}{
\fbox{\includegraphics[width=4.2in]{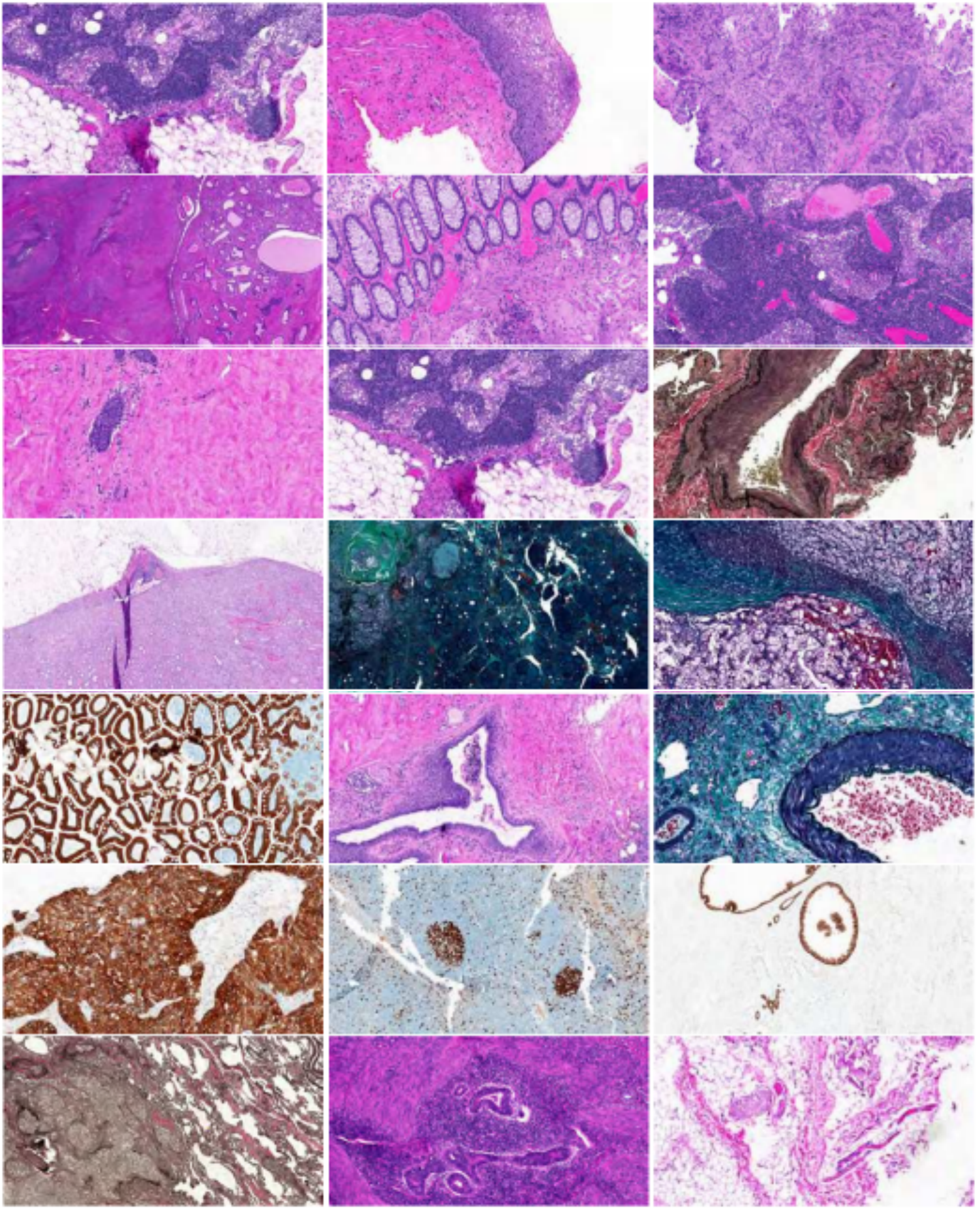}}
}
}
\caption{This figure shows all 24 whole slide images used to generate the KimiaPath24 dataset\cite{babaie2017classification}.
}
\label{fig:kimia24}
\end{figure}

Conversely, it is relatively easy to obtain weakly labeled images as it does not require annotation from medical experts. One example of such a dataset is KimiaPath24~\cite{babaie2017classification} (see Figure~\ref{fig:kimia24}). These images are collected from different parts of the body by non-medical experts based on visual similarity and are organized into 24 tissue groups.

In this work, our goal is to transfer knowledge from weakly labeled data to a dataset where labeled data is scarce.
We compare three approaches. As a baseline, we train on the target dataset from scratch, i.e., from a randomly initialized model. We contrast this with models pre-trained on weakly labeled data by using two strategies:  First, we learn features via classification of groups with a standard cross-entropy loss. Second, we explore metric learning as an alternative method for feature learning.
In order to evaluate our method, we simulate data scarcity by varying the amount of data available from the richly annotated CRC~\cite{kather2016multi} and PCam~\cite{veeling2018rotation} datasets.

Our contributions are as follows. First, we show that features learned from weakly labeled data are indeed useful for training models on other histology datasets.
Second, we show that with weakly labeled data, one can use an order of magnitude less data to achieve competitive patch classification results rivaling models trained from scratch with 100\% of the data.
Third, we further explore a proxy-based metric learning approach to learn features on the weakly labeled dataset. Finally, we also achieve the state-of-the-art results for both CRC and PCam when our models are trained with both weakly labeled data and 100\% of the original annotated data.\looseness=-1

\section{Related Work}\label{sec:related}

Recently in the field of DP, there are a few works that try to perform transfer learning from one dataset to another. Khan et al.~\cite{khan2019improving} attempt to improve prostate cancer detection by pre-training on a breast cancer dataset. Medela et al.~\cite{medela2019few} perform few-shot learning for lung, breast, and colon tissues by pre-training on the CRC dataset. These works demonstrate that there are some transferable features among various organs. However, the source domain on which they pre-train the model are still annotated by medical experts.

Apart from using classification to learn features from weakly labeled data, we explore the use of metric learning as an alternative approach. Metric Learning is a machine learning task where we learn a function to capture similarity or dissimilarity between inputs.
Metric Learning has many applications such as person re-identification~\cite{luo2019bag}, product retrieval~\cite{oh2016deep}, clothing  retrieval~\cite{liu2016deepfashion} and face recognition~\cite{wen2016discriminative}. In Digital Pathology, Medela et al.~\cite{medela2019few} use metric learning in the form of Siamese networks for few-shot learning tasks. Teh et al.~\cite{teh:MIDLAbstract2019a} also use Siamese networks to boost patch classification performance in conjunction with cross-entropy loss. In Siamese networks, pairs of images are fed to the network, and the network attracts or repels these images based on some established notion of similarity (e.g., class label). A shortcoming for Siamese networks is its sampling process, which grows quadratically with respect to the number of examples in the dataset \cite{movshovitz2017no, wu2017sampling,wang2019multi}. Due to this shortcoming, we choose to explore ProxyNCA~\cite{movshovitz2017no}, which offers faster convergence and better performance.

\section{Description of datasets}\label{sec:method}

To validate our hypothesis, we use one weakly labeled dataset and two target datasets where patch classification is the targeted task. First, we pre-train two different models (Classification and ProxyNCA) on the weakly labeled dataset. After pre-training, we train and evaluate our models on the target datasets.

\subsection{Weakly labeled dataset}

We use the KimiaPath24 dataset \cite{babaie2017classification} as our weakly labeled dataset, where we learn transferable features. A medical non-expert has collected this data from various parts of the body and has organized them into 24 groups based on visual distinction. These images have a resolution of \SI{0.5}{\micro\metre} per pixel with a size of $1000 \times 1000$ pixels. There are a total of 23,916 images in this dataset.\\

\subsection{Target Dataset A: Colorectal Cancer (CRC) Dataset}
The CRC dataset\cite{kather2016multi} consists of seven types of cell textures: tumor epithelium, simple stroma, complex stroma, immune cells, debris and mucus, mucosal glands, adipose tissue, and background. There are 625 images per class, and each image has a resolution of \SI{0.49}{\micro\metre} per pixel with dimensions $150 \times 150$ pixels.

\begin{figure}[h]
\centering
{\setlength{\tabcolsep}{2pt}
\scalebox{0.73}{
\fbox{
\begin{tabular}{cccc}
\includegraphics[width=1.0in,height=1.0in]{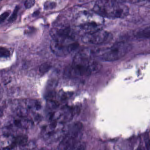}&
\includegraphics[width=1.0in,height=1.0in]{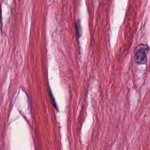}&
\includegraphics[width=1.0in,height=1.0in]{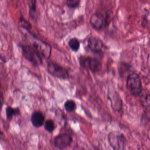}&
\includegraphics[width=1.0in,height=1.0in]{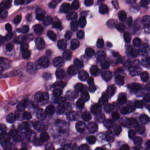} \\
Tumour Epithelium & Simple Stroma & Complex Stroma & Immune Cells\\
 & & &\\
\includegraphics[width=1.0in,height=1.0in]{figs/colon/05.png}&
\includegraphics[width=1.0in,height=1.0in]{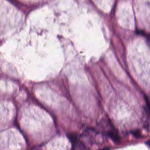}&
\includegraphics[width=1.0in,height=1.0in]{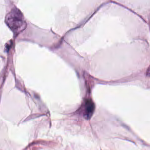}&
\includegraphics[width=1.0in,height=1.0in]{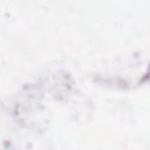}\\
Debris and Mucus & Mucosal Glands & Adipose Tissue & Background\\
\end{tabular}
}
}
}
\caption{Examples of tissue samples from the CRC dataset}
\label{fig:colon}
\end{figure}

\begin{figure}[h]
\centering
{\setlength{\tabcolsep}{2pt}
\scalebox{0.73}{
\fbox{
\begin{tabular}{ccccc}
\parbox[t]{3mm}{\multirow{1}{*}{\rotatebox[x=-10in]{90}{\vspace{10pt}\textbf{Normal}\hspace{-40pt}}}}&
\includegraphics[width=1.0in,height=1.0in]{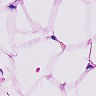}&
\includegraphics[width=1.0in,height=1.0in]{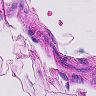}&
\includegraphics[width=1.0in,height=1.0in]{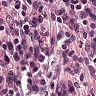}&
\includegraphics[width=1.0in,height=1.0in]{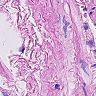} \\
\parbox[t]{3mm}{\multirow{1}{*}{\rotatebox[x=-10in]{90}{\vspace{10pt}\textbf{Tumor}\hspace{-40pt}}}}&
\includegraphics[width=1.0in,height=1.0in]{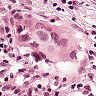}&
\includegraphics[width=1.0in,height=1.0in]{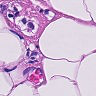}&
\includegraphics[width=1.0in,height=1.0in]{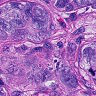}&
\includegraphics[width=1.0in,height=1.0in]{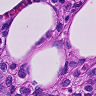}\\
\end{tabular}
}
}
}
\caption{Examples of tissue samples from the PCam dataset.
}

\label{fig:pcam}
\end{figure}

\subsection{Target Dataset B: Patch Camelyon (PCam) Dataset}
PCam~\cite{veeling2018rotation} is a subset of the CAMELYON16 dataset~\cite{bejnordi2017diagnostic}, which is a breast cancer dataset. There are a total of 327,680 images with a resolution of \SI{0.97}{\micro\metre} per pixel and dimensions $96 \times 96$ pixels. There are two categories presented in this dataset: tumor and normal, where an equal number of patches are present in both categories.

\section{Deep metric learning}

We propose to use ProxyNCA (Proxy-Neighborhood Component Analysis)~\cite{movshovitz2017no}, which is a proxy-based metric learning technique.
One benefit of ProxyNCA over standard Siamese Networks is a reduction in the number of compared examples, which is achieved by comparing examples against the class proxies.

\begin{figure}[H]
\centering
{\setlength{\tabcolsep}{2pt}
\scalebox{0.60}{
\begin{tabular}{cccc}
\fbox{\includegraphics[width=2.5in,height=1.3in]{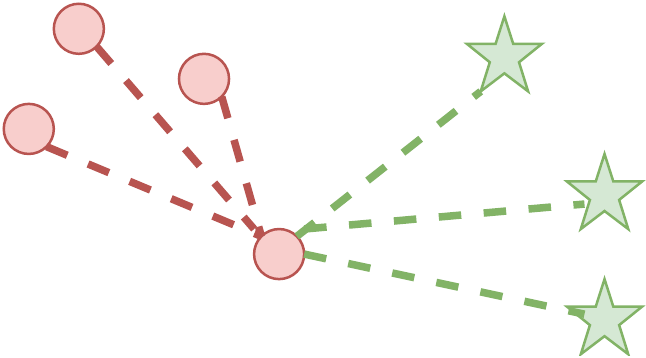}}&
\fbox{\includegraphics[width=2.5in,height=1.3in]{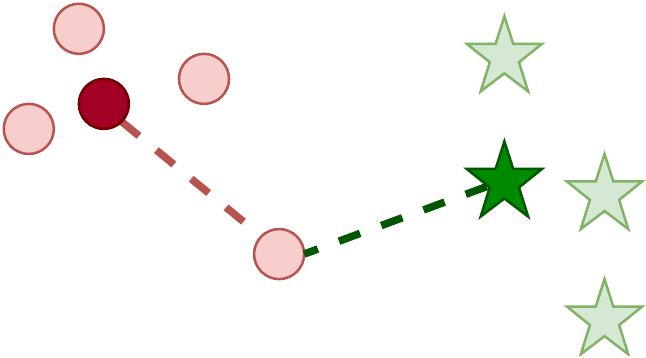}}&\\
\end{tabular}
}
}
\caption{A visualization how ProxyNCA works. [Left panel] Standard NCA compares one example with respect to all other examples (8 different pairings). [Right panel] In ProxyNCA, we only compare to the class proxies (2 different pairings) The above images are reproduced from~\cite{movshovitz2017no}.
}
\label{fig:pnca}
\end{figure}

Figure~\ref{fig:pnca} visualizes the computational savings that can be gained by comparing examples to class proxies rather than other examples. In ProxyNCA, class proxies are stored as parameters with the same dimension as the input embedding space and are updated by minimizing the proxy loss $l$.

ProxyNCA requires an input example $x_i$, a backbone Model $M$, an embedding Layer $E$, a corresponding embedding $\alpha_i$, a proxy function $p(\alpha_i)$, which returns a proxy of the same class as $\alpha_i$, and a proxy function $z(\alpha_i)$, which returns all proxies of different class than $\alpha_i$. Its goal is to minimize the distance between $\alpha_i$ and $p(\alpha_i)$ and at the same time maximize the distance between $\alpha_i$ and $z(\alpha_i)$. Let $d(a,b)$ denote Euclidean distance between $a$ and $b$, ${||a||}_2$ be the $L^2$-Norm of vector $a$ and $\lambda$ be the learning rate. We describe how to train ProxyNCA in Algorithm \ref{alg:proxynca}.

\begin{algorithm}
   \caption{ProxyNCA Training}
         \label{alg:proxynca}
    \begin{algorithmic}[1]
    \State Randomly initialize all proxies
    \For{$j = 1 \ldots $ number of mini-batches}
        \For{$i = 1 \ldots $ mini-batch size}
            \State $\alpha_i = E(M(x_i))$
            \State $l = - \log \left(\frac{\exp^{-d\left(\frac{\alpha_i}{{||\alpha_i||}_2}, \frac{p\left(\alpha_i\right)}{{||p\left(\alpha_i\right)||}_2}\right)}}
                                          {\sum_{z \in Z(\alpha_i)} \exp^{-d\left(\frac{\alpha_i}{{||\alpha_i||}_2}, \frac{z}{{||z||}_2}\right)}}\right)$
            \State $\theta = \theta - \lambda \frac{\partial l}{\partial \theta}$
        \EndFor
    \EndFor
\end{algorithmic}
\end{algorithm}

After training on weakly labeled data, we discard everything except for the backbone model. We then use this backbone model along with a new embedding layer to train on the target dataset.

\section{Experiments}\label{sec:experiment}

In this section, we describe our experimental setup and results. We evaluate our models using prediction accuracy. For the colon cancer dataset, we follow the same experimental setup as \cite{kather2016multi} where we train and evaluate our models using 10-fold cross-validation. For the PCam dataset~\cite{veeling2018rotation}, we train our model using images in the training set and evaluate our performance on the test set following the setup from \cite{veeling2018rotation} and \cite{teh:MIDLAbstract2019a} where we report on different amounts ($R\%$) of labeled target data used in fine-tuning. $R\%$ of data is sampled evenly for all classes in the target dataset.

\begin{table}[H]
\centering
\setlength{\tabcolsep}{3pt}
\begin{tabular}{cccc} \hline
$N_c$ & $R$\% & Initialization/Method & Accuracy (\%) \\ \hline
12 & 2 & Random & 61.62 $\pm$ 3.79\\
  &  & ImageNet & 79.16 $\pm$ 2.83\\
  &  & Classification & \textbf{83.30 $\pm$ 2.88}\\
  &  & ProxyNCA & 82.82 $\pm$ 2.70\\  \hline
25 & 4 & Random & 64.12 $\pm$ 3.73\\
 &  & ImageNet & 83.42 $\pm$ 1.57\\
 &  & Classification & \textbf{87.30 $\pm$ 1.40}\\
 &  & ProxyNCA & 87.10 $\pm$ 1.23\\  \hline
50 & 9 & Random & 77.34 $\pm$ 2.11\\
 &  & ImageNet & 88.06 $\pm$ 1.73\\
 &  & Classification & 89.38 $\pm$ 1.29\\
 &  & ProxyNCA & \textbf{89.80 $\pm$ 1.88}\\  \hline
100 & 18 & Random & 84.58 $\pm$ 1.19\\
 &  & ImageNet & 90.08 $\pm$ 1.72\\
 &  & Classification & 91.58 $\pm$ 1.08\\
 &  & ProxyNCA & \textbf{91.96 $\pm$ 0.99}\\ \hline
625 & 100 & Random & 89.84 $\pm$ 1.29\\
 &  & ImageNet & 91.62 $\pm$ 1.38\\
 &  & Classification & 92.10 $\pm$ 0.80\\
 &  & ProxyNCA & \textbf{92.46 $\pm$ 1.22}\\
 &  & RBF-SVM \cite{kather2016multi}& 87.40 \\
\hline
\end{tabular}
\bigskip
    \caption{Accuracy of our model trained with $R$\% of data in three different pre-trained settings on the CRC dataset. $N_c$ denotes the number of examples per class.  We compare our approach with  the best model in \cite{kather2016multi} which is a RBF-SVM that uses five different concatenated features (Lower-order histogram, Higher-order histogram, Local Binary Patterns, Gray-Level Co-occurrence Matrix and Ensemble of Decision Trees).}
    \label{table:exp_1}
\end{table}

\subsection{Experimental Setup}

In all of our experiments, we use the Adam~\cite{kingma2014adam} optimizer to train our model. Following \cite{movshovitz2017no}, we use an exponential learning rate decay schedule with a factor of 0.94. We also perform channel-wise data normalization with the mean and standard deviation of the respective dataset.  \\

\noindent{\textbf{Weakly labeled dataset:}}
We train our weakly labeled dataset similarly for both target datasets with one subtle difference. For the CRC dataset, we perform random cropping of size $150 \times 150$ pixels directly on KimiaPath24 images since the resolution of both datasets are about the same. However, for the PCam dataset, we resize the KimiaPath24 images from $1000 \times 1000$ pixels to $514 \times 514$ pixels to match the resolution of the target dataset. Finally, we perform random cropping of size $96 \times 96$ pixels on the resized images. The initial learning rate is set to $1e^{-4}$, and we train our models for a fixed 100 epochs.\\

\noindent{\textbf{Target dataset:}} During training, we pad the images by 12.5\% via image reflection prior to random cropping to preserve the image resolution. We set the learning rate to $1e^{-4}$ and train all of our models with a fixed number of epochs (200 for the CRC dataset and 100 for the PCam dataset). We repeat the experiment ten times with different random seeds and report the mean over trials in Table~\ref{table:exp_1} and \ref{table:exp_2}. For the experiments that involve pre-training (Classification and ProxyNCA), we initialize our model with weights pre-trained on the weakly labeled dataset.   \\

\noindent{\textbf{Data augmentation:}} In addition to random cropping, we also perform the following data augmentation at each training stage: a) Random Horizontal Flip b) Random Rotation c) Color Jittering - where we set the hue, saturation, brightness and contrast threshold to 0.4.  All data transformations are implemented via the TorchVision package. \\

\noindent{\textbf{Backbone Model:}} In all of our experiments, we use a modified version of ResNet34~\cite{he2016deep}. Due to low resolution of the target datasets ($150 \times 150$ px and $96 \times 96$ px), we remove the max-pooling layer from ResNet34. \\

\noindent{\textbf{Additional Info:}} For the experiment with 100\% of training data on the PCam dataset, we follow the same experimental setup as \cite{teh:MIDLAbstract2019a} where the number of training epochs is ten and learning rate is reduced by a factor of ten after epoch five.

\subsection{Results}
Table~\ref{table:exp_1} and \ref{table:exp_2} show our experimental results. We train our model by varying the amount of target domain data on four different pre-training settings (Random, ImageNet, Classification and ProxyNCA). With the ImageNet setting, we initialize the model with weights pre-trained on the ImageNet 2012 dataset~\cite{deng2009imagenet}.

By pre-training our models on weakly labeled data, we achieve test accuracies of $89.80\%$ and $89.77\%$ on the CRC and PCam datasets respectively with an order of magnitude less training data. Both results rival the test accuracies of randomly initialized models~(89.84\% and 88.98\%), which are trained with $100\%$ of the data. With 100\% of training data, our models attain test accuracies of $92.70\%$ and $90.47\%$, outperforming previous state-of-the-art results: $87.40\%$ and $90.36\%$ on CRC and PCam respectively. We note that for PCam, the previous SOTA falls within the error bars of ours.

In addition, we also show that models pre-trained on weakly labeled data outperform models pre-trained on the ImageNet 2012 dataset, especially in the lower data regime. In Table~\ref{table:exp_2}, we observe that the performance of models trained under ImageNet setting underperform the randomly initialized model when $R$ is $10\%$ and $100\%$. We speculate that as we collect more images per class, the advantage of the ImageNet pre-trained model gradually vanishes.
This trend is not surprising due to the vast differences in the domain between two datasets (natural images vs.~Digital Pathology images).

We further observe that ProxyNCA outperforms classification on the PCam dataset. However, on the CRC dataset, this trend persists when the number of classes per sample is 50 or larger.
When the number of samples per class is too small, the results are highly varied, which makes the comparison more difficult.

\section{Conclusion}\label{sec:conclude}

We show that useful features can be learned from weakly labeled data, where the images are collected from different parts of the body by non-medical experts based on visual similarity. We show that such features are transferable to the CRC dataset as well as the PCam dataset and are able to achieve competitive results with an order of magnitude less training data. Although evaluation is facilitated in the simulated ``low data'' regime, our approach holds promise for transfer to digital pathology datasets for which the number of actual annotations by medical experts is very small.

\begin{table}[h]
\centering
\setlength{\tabcolsep}{3pt}
\begin{tabular}{cccc} \hline
$N_c$ & $R$\% & Initialization/Method & Accuracy (\%) \\ \hline
1,000 & 0.76 & Random & 79.37 $\pm$ 1.33 \\
  &  & ImageNet & 83.91 $\pm$ 0.84\\
 & & Classification & 85.29 $\pm$ 1.08\\
 & & ProxyNCA & \textbf{86.69 $\pm$ 0.46}\\  \hline
2,000 & 1.53 & Random & 81.26 $\pm$ 1.47\\
  &  & ImageNet & 86.12 $\pm$ 0.60\\
 & & Classification & 86.55 $\pm$ 1.29\\
 & & ProxyNCA & \textbf{87.38 $\pm$ 0.78}\\  \hline
3,000 & 2.29 & Random & 84.09 $\pm$ 1.24\\
  &  & ImageNet & 86.52 $\pm$ 0.63\\
 &  & Classification & 87.03 $\pm$ 0.62\\
 &  & ProxyNCA & \textbf{87.69 $\pm$ 0.72}\\ \hline
13,107 & 10 & Random & 88.47 $\pm$ 0.59\\
  &  & ImageNet & 87.22 $\pm$ 0.47\\
 &  & Classification & 89.64 $\pm$ 0.39\\
 &  & ProxyNCA & \textbf{89.77 $\pm$ 0.50}\\ \hline
131,072 & 100 & Random & 88.98 $\pm$ 1.06\\
 &  & ImageNet & 88.91 $\pm$ 0.53\\
 &  & Classification & 89.85 $\pm$ 0.64\\
 &  & ProxyNCA & \textbf{90.47 $\pm$ 0.59}\\
 &  & P4M\cite{veeling2018rotation} & 89.97\\
 &  & Pi+\cite{teh:MIDLAbstract2019a} & 90.36 $\pm$ 0.41\\
\hline
\end{tabular}
\bigskip
    \caption{Accuracy of our model trained with $R$\% of data in three different pre-trained settings on the PCam dataset. $N_c$ denotes the number of examples per class. We compare our approach to \cite{veeling2018rotation}, which uses a DenseNet with group convolutional layers and \cite{teh:MIDLAbstract2019a} which uses contrastive and self-perturbation loss together with cross entropy loss.
    }
    \label{table:exp_2}
\end{table}

\bibliographystyle{IEEEbib}
\bibliography{isbi}

\end{document}